\documentclass[runningheads]{llncs}

\usepackage{eccv}

\usepackage{eccvabbrv}

\usepackage{graphicx}
\usepackage{booktabs}

\usepackage[accsupp]{axessibility}  

\usepackage{hyperref}

\usepackage{orcidlink}

\usepackage{makecell}
\usepackage{multirow}

\usepackage{color}
\usepackage{adjustbox}
\usepackage{array}
\usepackage{xcolor,colortbl}
\usepackage{balance}

\newcommand{\Amat}{{\bf A}}

\newcommand{\Mmat}[0]{{{\bf M}}}

\newcommand{\Xmat}{{\bf X}}
\newcommand{\Ymat}[0]{{{\bf Y}}}

\newcommand{\kv}[0]{{\boldsymbol{k}}}

\newcommand{\qv}[0]{{\boldsymbol{q}}}

\newcommand{\vv}{\boldsymbol{v}}

\usepackage{marvosym}

\begin{document}

\title{A Simple Low-bit Quantization Framework for Video Snapshot Compressive Imaging} 

\titlerunning{A Simple Low-bit Quantization Framework for Video SCI}

\author{Miao Cao\inst{1,2} \and
Lishun Wang\inst{2}\and
Huan Wang\inst{2}\and
Xin Yuan\inst{2}\textsuperscript{(\Letter)}}

\authorrunning{M. Cao et al.}

\institute{Zhejiang University, Hangzhou 310058, Zhejiang, China \and
School of Engineering, Westlake University, Hangzhou, 310030, Zhejiang, China\\
\email{\{caomiao,wanglishun,wanghuan,xyuan\}@westlake.edu.cn}}

\maketitle

\begin{abstract}
  Video Snapshot Compressive Imaging (SCI) aims to use a low-speed 2D camera to capture high-speed scene as snapshot compressed measurements, followed by a reconstruction algorithm to reconstruct the high-speed video frames.  
  State-of-the-art (SOTA) deep learning-based algorithms have achieved impressive performance, yet with heavy computational workload. 
  Network quantization is a promising way to reduce computational cost. However, a direct low-bit quantization will bring large performance drop. 
  To address this challenge, in this paper, we propose a simple low-bit quantization framework ({dubbed~\em Q-SCI}) for the end-to-end deep learning-based video SCI reconstruction methods which usually consist of a feature extraction, feature enhancement, and video reconstruction module. 
  Specifically, we first design a high-quality feature extraction module and a precise video reconstruction module to extract and propagate high-quality features in the low-bit quantized model.
  In addition, to alleviate the information distortion of the Transformer branch in the quantized feature enhancement module, we introduce a shift operation on the query and key distributions to further bridge the performance gap. Comprehensive experimental results manifest that our Q-SCI framework can achieve superior performance, \eg, 4-bit quantized EfficientSCI-S derived by our Q-SCI framework can theoretically accelerate the real-valued EfficientSCI-S by 7.8$\times$ with only 2.3$\%$ performance gap on the simulation testing datasets. Code is available at \url{https://github.com/mcao92/QuantizedSCI}. 
\end{abstract}

\section{\bf Introduction}

\begin{figure}[!t]
    \centering
    \begin{minipage}{0.55\linewidth}
        \centering
        \includegraphics[width=\linewidth]{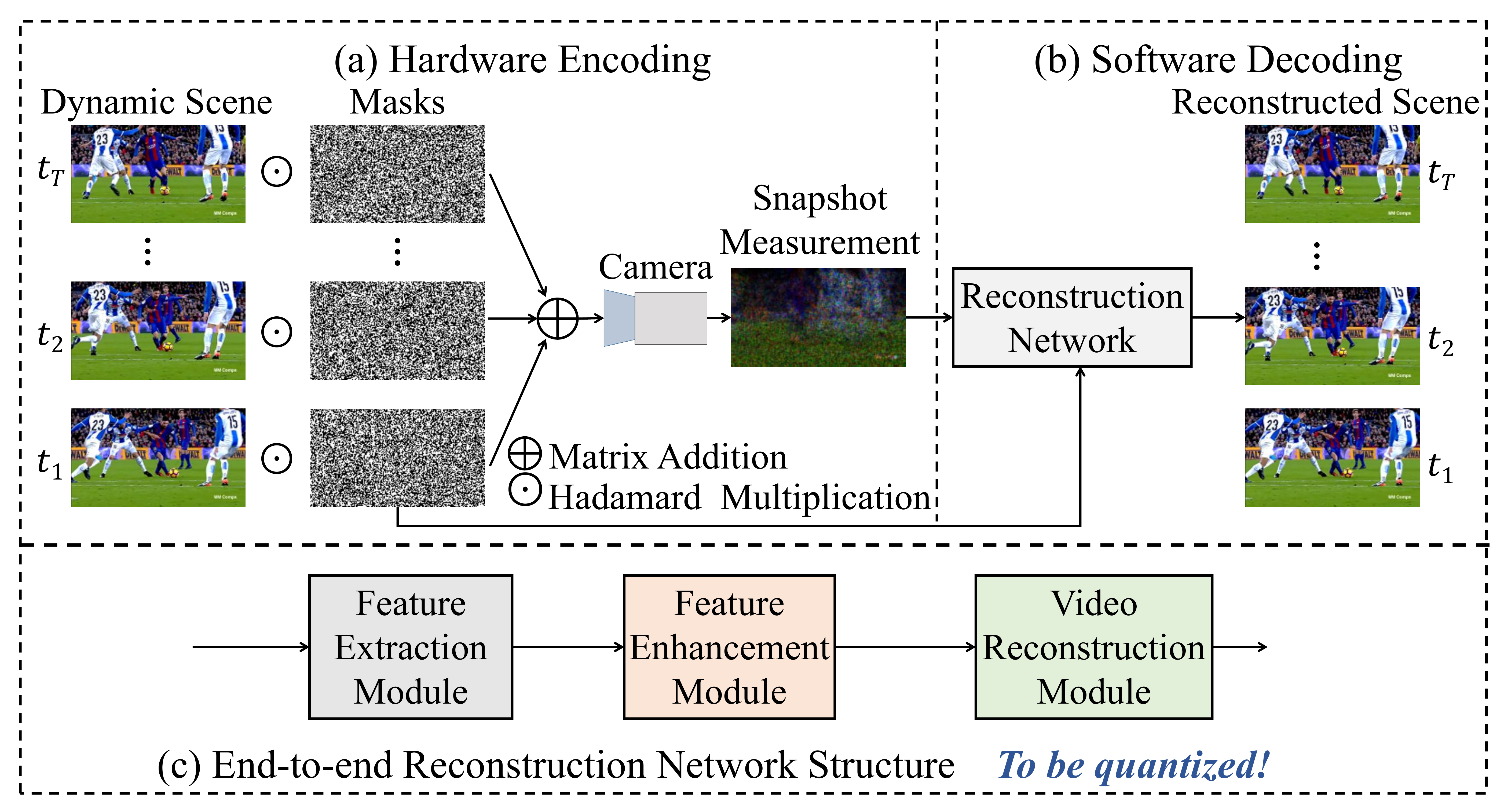}
    \end{minipage}\hfill
    \begin{minipage}{0.4\linewidth}
        \centering
        \includegraphics[width=\linewidth]{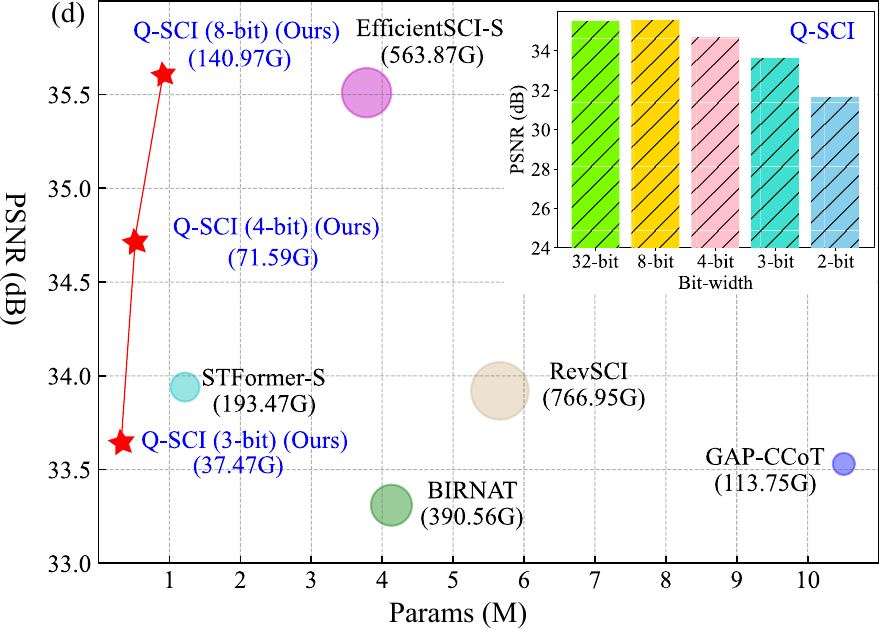}
    \end{minipage}
    \caption{\small{
        (a) In the encoding process of video SCI, a dynamic scene is modulated by different masks and then the modulated scene is captured by a camera as snapshot measurements. (b) In the decoding process of video SCI, the captured measurements and the corresponding masks are fed into a reconstruction algorithm to obtain the desired video frames. (c) The pipeline of an end-to-end video SCI reconstruction network. (d) Comparison of the reconstruction quality and computational cost of different reconstruction methods. 
        }} 
    \label{fig:para_sota}
\end{figure}

Recently, video Snapshot Compressive Imaging (SCI) has attracted much attention because it can capture high-speed scenes using a low-speed 2D camera with low bandwidth. There are two main steps in the video SCI system: hardware encoding (shown in Fig.~\ref{fig:para_sota}(a)) and software decoding(shown in Fig.~\ref{fig:para_sota}(b))~\cite{yuan2021snapshot}. In the hardware encoding process, we first modulate the dynamic scene with different masks, and then the modulated scene is compressed into a series of snapshot measurements, which are finally captured by a low-cost 2D camera. In the software decoding stage, the captured snapshot measurements and the corresponding modulation masks are fed into a reconstruction algorithm to recover the desired video frames. 

On the one hand, many successful video SCI hardware encoders~\cite {reddy2011p2c2,hitomi2011video,deng2019sinusoidal,zhang2021ten} have been built. On the other hand, state-of-the-art (SOTA) reconstruction algorithms~\cite {cheng2022recurrent,cheng2021memory,wang2021metasci,yang2022ensemble,wu2021dense,wang2022snapshot}, mostly based on deep neural networks, have outperformed the traditional model-based methods~\cite {liu2018rank,yuan2016generalized} by a large margin not only in the reconstruction quality but also in the running speed. Thus, it seems that the current video SCI systems can already be used in practical applications. Unfortunately, existing deep learning-based reconstruction algorithms usually need a large number of parameters and floating point operations (FLOPs) to achieve satisfactory accuracy. EfficientSCI~\cite{wang2023efficientsci} is a significant exploration to design an efficient video SCI reconstruction algorithm, which has 3.78 MB parameters and requires 563.87 GFLOPs to reconstruct the video frames with 35.51 dB on the simulation testing datasets. Our ultimate goal is to deploy video SCI reconstruction methods on AI chips, thus to forming a real end-to-end system integrating capture and reconstruction. However, it is still challenging to deploy previous reconstruction algorithms on resource-limited devices due to their high computational cost. This motivates us to move one step further to reduce the computational workload of the video SCI reconstruction algorithms while preserving the model performance as much as possible.

Substantial efforts have been made to compress and accelerate neural networks for efficient online inference. These methods can be divided into the following categories: network quantization~\cite{yuan2022ptq4vit,yao2022zeroquant}, network pruning~\cite{wang2021neural,li2022revisiting,fang2023depgraph}, knowledge distillation~\cite{wang2020collaborative,gou2021knowledge,wang2022r2l,zhao2022decoupled}, and compact network design~\cite{chen2022mobile,li2022efficientformer,li2023rethinking}. Among them, network quantization is suitable for deployment on resource-limited platforms because it can reduce the bit-width of network parameters and
activations for efficient inference. However, directly applying low-bit quantization to video SCI reconstruction algorithms may encounter the following issues: {\bf {i)}} Directly quantizing previous reconstruction methods into low-bit will lead to a large performance drop from their full-precision counterpart. {\bf  {ii)}} There exists a distortion of the query and key distributions in the low-bit quantized vision Transformer.    

Bearing the above concerns in mind, in this paper, we propose a simple low-bit quantization framework, dubbed {\em{Q-SCI}}, for the end-to-end deep learning-based video SCI reconstruction methods which usually consist of a feature extraction module, a feature enhancement module, and a video reconstruction module as shown in Fig.~\ref{fig:para_sota}(c). Specifically, we first conduct extensive empirical analysis and identify that the large performance drop in low-bit quantized model mainly comes from the information loss of low-bit quantized features. Therefore, we design a high-quality feature extraction module and a precise video reconstruction module which can extract and propagate high-quality features through the low-bit quantized network. In addition, to alleviate the information distortion of the Transformer branch in the quantized feature enhancement module, we introduce a shift operation on the query and key distributions to further bridge the performance gap. 
Finally, following previous works~\cite{wang2022learnable,ma2023ompq,liu2022nonuniform} which are built and evaluated on existing backbones. We choose EfficientSCI-S as the full-precision backbone model for our proposed Q-SCI because EfficientSCI-S can achieve SOTA reconstruction quality with much cheaper computational cost. As shown in Fig.~\ref{fig:para_sota}(d), compared with previous methods, the proposed Q-SCI framework can achieve comparable reconstruction quality with {\em much fewer model parameters and FLOPs}.

The main contributions of our paper can be summarized as follows:  
\begin{itemize}
    \item We propose a simple low-bit quantization framework for video SCI reconstruction. 
    To our best knowledge, this is {\em the first network quantization framework} in the video SCI reconstruction task.
    \item We design {\em a high-quality feature extraction module and a precise video reconstruction module} to better extract and propagate high-quality features. Furthermore, the {\em generalization ability} of the proposed Q-SCI framework is verified on different end-to-end video SCI reconstruction methods.  
    \item {\em A shift operation} on the query and key distributions is introduced to alleviate the information distortion in the quantized Transformer branch.
    \item Experimental results on the simulated and real datasets demonstrate that the low-bit quantized models derived by our proposed Q-SCI framework can achieve superior performance with {\em much cheaper computational cost}. 
\end{itemize}

\section{\bf Related Work}
\label{Sec:related}

\subsection{\bf Video SCI Reconstruction Algorithms}
The existing video SCI reconstruction methods can be divided into two categories: {\em traditional model-based methods} and {\em deep learning-based ones}.
The traditional model-based methods formulate the video SCI reconstruction process as an optimization problem with a regularization term such as total variation~\cite{yuan2016generalized} and Gaussian mixture model~\cite{yang2014compressive}, and solve it via iterative algorithms. The major drawback of these model-based methods is that one needs to perform time-consuming iterative optimization. 
In order to improve the running speed, Yuan {\em et al.} develop a plug-and-play (PnP) framework, in which a pre-trained denoising model is plugged into every iteration of the optimization process~\cite{yuan2021plug,yuan2020plug}. However, they still take a long time on reconstructing large-scale data. 

Recently, more and more researchers begin to utilize deep neural networks in the video SCI reconstruction task.
For example, BIRNAT~\cite{cheng2022recurrent} uses a bidirectional recurrent neural network to exploit the temporal correlation.
To save model training memory, RevSCI~\cite{cheng2021memory} builds an end-to-end 3D convolutional neural network with reversible structure for video SCI reconstruction. More recently, Wang {\em et al.} build the first Transformer-based reconstruction network~\cite{wang2022spatial} with space-time factorization and local self-attention mechanism. After that, Wang {\em et al.} design an efficient reconstruction network~\cite{wang2023efficientsci,cao2024hybrid} based on dense connections and space-time factorization.
Combining the idea of model-based and deep learning-based methods, researchers propose the deep unfolding networks~\cite{wu2021dense,yang2022ensemble,zheng2023unfolding}. 
For example, Zheng {\em et al.} first introduce the uncertainty estimation mechanism into the deep unfolding framework~\cite{zheng2023unfolding}.  
Although the deep learning-based models have achieved impressive results, it is still challenging to deploy them on the resource-limited devices due to their high computational workload. In this paper, we mainly focus on developing {\em light-weight video SCI reconstruction algorithm} empowered by network quantization.  

\subsection{\bf Network Quantization} 
Due to its impressive effectiveness in computational compression, network quantization has
been widely applied in high-level vision~\cite{wang2020towards,chen2019deep,xu2020generative,li2023q} and low-level vision tasks~\cite{li2020pams,hong2022cadyq,xia2022basic,cai2023binarized,qin2024quantsr}. {\em In the high-level vision tasks}, Liu {\em et al.} develop a dynamic quantization scheme~\cite{liu2022instance} to decide the optimal bit-widths for each image. GPUSQ-ViT~\cite{yu2023boost} is a compression scheme which maximally utilize the GPU-friendly 2:4 fine-grained structured sparsity and quantization. Q-ViT~\cite{li2022q} presents the first quantization-aware training framework for accurate and low-bit vision Transformer. ReActNet~\cite{liu2020reactnet} learns the distribution shape and shift to improve the performance of the CNN-based binary network. Li {\em et al.} propose a novel post-training quantization framework specifically tailored towards the unique multi-timestep pipeline and model architecture of the diffusion models~\cite{li2023q}. Li {\em et al.} develop a fully quantized network for object detection~\cite{li2019fully}. {\em Considering the low-level vision tasks}, PAMS~\cite{li2020pams} learns the quantization intervals of different layers to adapt to vastly distinct feature distributions in the super-resolution (SR) networks. Hong {\em et al.} propose CADyQ~\cite{hong2022cadyq} to assign different bit-width for each patch and layer for image SR. Xia {\em et
al.} design a binarized convolution unit BBCU~\cite{xia2022basic} for image restoration-tasks, \eg, image denoising and JPEG
compression artifact reduction. FQSR~\cite{wang2021fully} jointly optimize efficiency and accuracy on the image SR task. Yet, the potential of network quantization for the video SCI reconstruction task has not been explored.

\section{Prerequisites of Network Quantization}

In this section, we briefly introduce the network quantization architecture used in our paper. First, given the activation $x$ and weight ${\bf w}$, we introduce a general asymmetric activation quantization and symmetric weight quantization scheme, expressed as follows:
\begin{eqnarray}
    Q_a(x) &=& \textstyle \lfloor \text{clip}\{(x-z)/\alpha_x,-Q_n^x,Q_p^x\} \rceil,\nonumber\\
    \hat{x} &=& \textstyle Q_a(x) \times \alpha_x + z,\nonumber\\
    Q_{{\bf w}}({\bf w}) &=& \textstyle \lfloor \text{clip}\{{\bf w}/\alpha_{{\bf w}},-Q_n^{{\bf w}},Q_p^{{\bf w}}\} \rceil,\nonumber\\
    \hat{{\bf w}} &=& \textstyle Q_{{\bf w}}({\bf w}) \times \alpha_{{\bf w}},
\end{eqnarray}
where $\text{clip}\{x,m_1,m_2\}$ constrains $x$ into $[m_1,m_2]$ by setting $x$ as $m_1$ when $x < m_1$ and $m_2$ when $x > m_2$. $\lfloor x \rceil$ rounds $x$ to the nearest integer. $\alpha$ is a scale factor, which can divide the entire range of the input into uniform partitions. $z$ is a zero-point, which can shift the quantized distribution into the a specific range in the asymmetric quantization. During training, $\alpha$ and $z$ are initialized as 1 and 0, and then been optimized as network parameters.  
Following this, if we quantize the activation $x$ to $a$ bits and weight ${\bf w}$ to $b$ bits, we have:
\begin{eqnarray}
    Q_n^x = 2^{a-1},Q_p^x = 2^{a-1}-1,\nonumber\\
    Q_n^{{\bf w}} = 2^{b-1},Q_p^{{\bf w}} = 2^{b-1}-1.
\end{eqnarray}

In this way, the forward and back propagation of network quantization can be formulated as $Q\text{-Linear}(x)=\hat{x}\cdot\hat{{\bf w}}=\alpha_x\alpha_{{\bf w}}((Q_a{(x)}+z/\alpha_x) \otimes Q_{{\bf w}}{({\bf w})}$ and Eq.~\eqref{eq:forward-backward} respectively.
\begin{equation}
\label{eq:forward-backward}
    \begin{aligned}
    &{\textstyle\frac{\partial \mathcal{J}}{\partial x} =  \frac{\partial \mathcal{J}}{\partial \hat{x}} \frac{\partial \hat{x}}{\partial x} = \left\{
    \begin{array}{cl}
     \frac{\partial \mathcal{J}}{\partial \hat{x}} &\operatorname{if} x \in [-Q_n^x, Q_p^x] \\
     0 &\operatorname{otherwise}\;\;\;\;\;\;\;\;\;\;\;  \\
    \end{array}
    \right.} ,  \\
    &{\textstyle\frac{\partial \mathcal{J}}{\partial {\bf w}} =  \frac{\partial \mathcal{J}}{\partial x} \frac{\partial x}{\partial \hat{\bf w}} \frac{\partial \hat{\bf w}}{\partial {\bf w}} = \left\{
    \begin{array}{cl}
     \frac{\partial \mathcal{J}}{\partial x} \frac{\partial x}{\partial \hat{\bf w}} &\operatorname{if} {\bf w} \in [-Q_n^{\bf w}, Q_p^{\bf w}] \\
     0 &\operatorname{otherwise}\;\;\;\;\;\;\;\;\;\;\;\;\;  \\
    \end{array}
    \right. }, 
    \end{aligned}
\end{equation}
where $\mathcal{J}$ denotes the loss function, $Q(\cdot)$ is used in the forward process while the straight-through estimator~\cite{bengio2013estimating} is adopted to retain the derivation of gradient in the backward propagation process, and $\otimes$ denotes the efficient bit-wise matrix multiplication operation. 

Recent methods~\cite{wang2022spatial,wang2023efficientsci} begin to explore the use of Transformer in the end-to-end video SCI reconstruction methods. Thus, we define the quantization process of the self-attention module in the following way. First, we denote the quantized computation on the query $\qv$, key $\kv$ and value $\vv$ as $\qv=Q\text{-Linear}_q{(x)}$, $\kv=Q\text{-Linear}_k{(x)}$, $\vv=Q\text{-Linear}_v{(x)}$, where $Q\text{-Linear}_q$, $Q\text{-Linear}_k$ and $Q\text{-Linear}_v$ represent the quantized linear layers for $\qv$, $\kv$ and $\vv$, respectively. Then, the attention weight can be expressed as:
\begin{align}
    \Amat&=\textstyle \frac{1}{\sqrt{d}}(Q_a{(\qv)}\otimes Q_a{(\kv)}^\top), \nonumber\\
    Q_\Amat&=Q_a{(\text{softmax}(\Amat))}.
\end{align}

\begin{figure*}[!ht]
    \centering 
    \includegraphics[width=1.\linewidth]{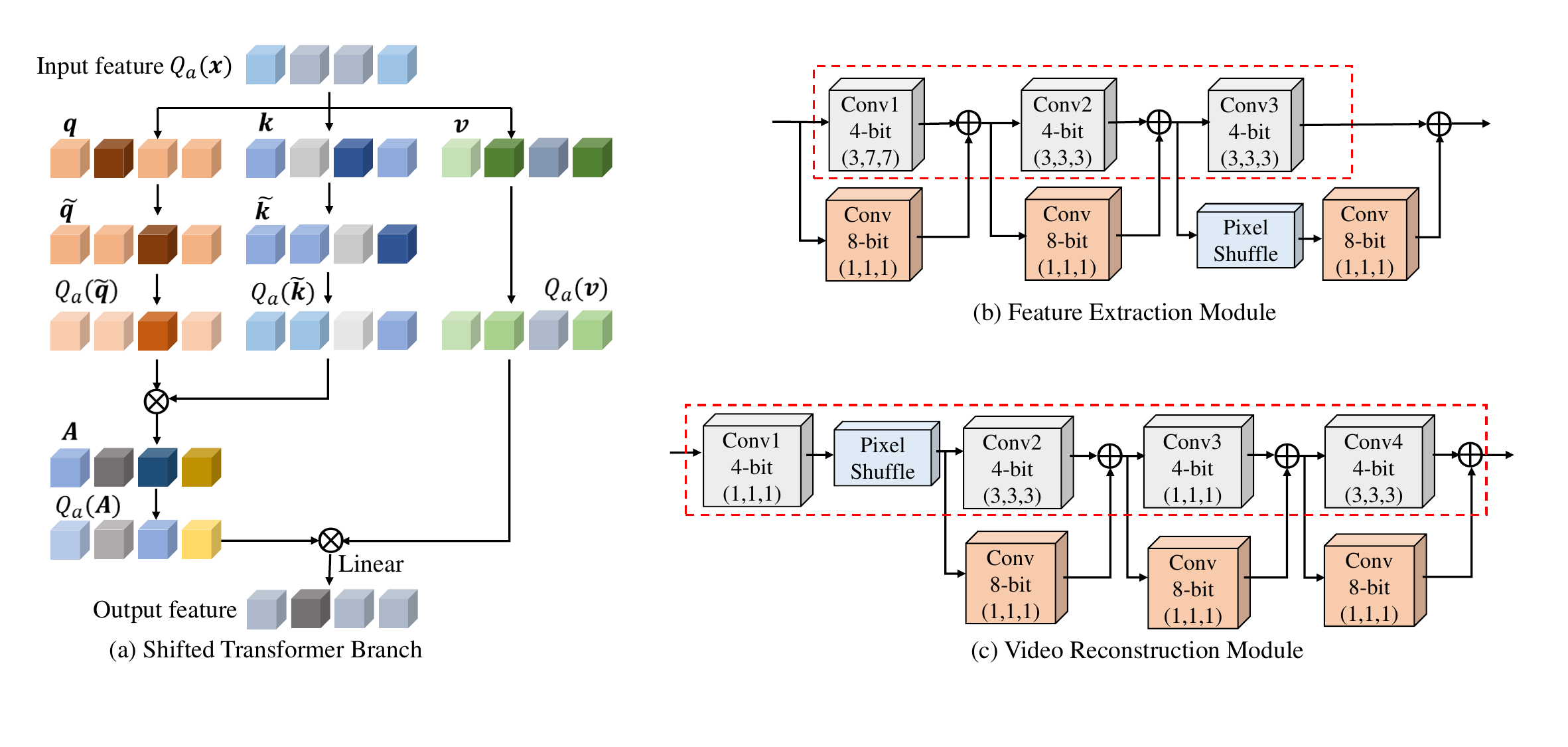}
    \caption{\small{Illustration of the proposed Q-SCI framework: In (a), 
    we introduce a {\em shift operation} on the query and key to alleviate their distribution distortion in the low-bit Transformer branch. In (b) and (c), convolutional layers in orange boxes are plugged to extract and propagate high-quality features through the low-bit quantized network.
    }}
  \label{fig:network}
\end{figure*}

\section{\bf Proposed Methods}
\label{Sec:net}

In this section, we propose an accurate and fully quantized framework for the end-to-end video SCI reconstruction algorithms. Firstly, let us give a brief introduction of the previous SOTA EfficientSCI which is used as the full-precision backbone model for our proposed Q-SCI framework. The ResDNet module with $N$ residual style ResDNet blocks is used for feature enhancement in EfficientSCI. In each ResDNet block, we put several CFormer blocks with {\em temporal Transformer branch}. Please refer to~\cite{wang2023efficientsci} and the supplementary material for more details about EfficientSCI and the video SCI mathematical model. Then, we conduct extensive empirical analysis on the severe performance drop of the low-bit quantized model. Following this, we design a high-quality feature extraction module, a precise video reconstruction module and a shifted Transformer branch to improve the performance of the low-bit quantized model. Afterwards, we customize four different quantized network variants based on our Q-SCI framework. Finally, the loss function is defined for the training process.    

\begin{table}[!ht]
  \setlength\tabcolsep{2pt}
  \caption{\small{Reconstruction quality of different models which directly quantizing the feature extraction module, ResDNet module and video reconstruction module of EfficientSCI-S into 4-bit.}}
  \centering
  \resizebox{.7\textwidth}{!}
  {
  \centering
  \begin{tabular}{c|c|c|c|c}
  \toprule
  \makecell[c]{Feature Extraction\\Module}
  & \makecell[c]{ResDNet\\Module}
  & \makecell[c]{Video Reconstruction\\Module} 
  & PSNR 
  & SSIM
  \\
  \midrule
  \rowcolor{lightgray}
  & & &35.23 &0.968
  \\
  \checkmark & & &33.01 &0.941
  \\
  & \checkmark & &34.71 &0.963
  \\
   & & \checkmark &34.74 &0.962
  \\
  \bottomrule
  \end{tabular}
  }
  \label{Tab:feat}
\end{table}

\subsection{\bf Performance Analysis on Low-bit Quantization}
\label{sec:analysis}

\begin{figure}[!t]
    \centering 
    \includegraphics[width=1.\linewidth]{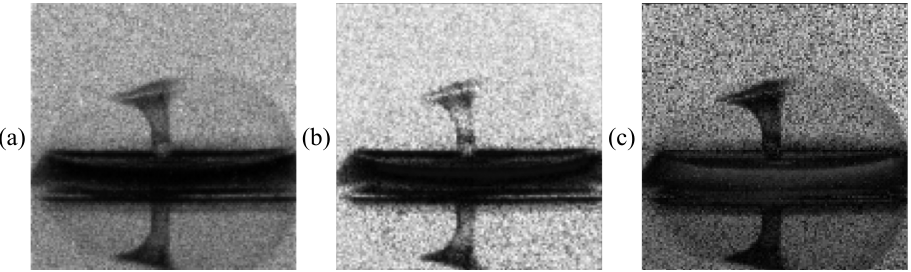}
    \caption{\small{The output feature map of the feature extraction module under different models: (a) Full-precision model, (b) 4-bit quantized baseline model, and (c) Using the proposed high-quality feature extraction module in 4-bit quantized baseline model.} 
    }
  \label{fig:feat_map}
\end{figure}

In the experiments, compared with the full-precision backbone model EfficientSCI-S, we observe a large performance drop (4.11 dB) in the 4-bit quantized baseline model which directly quantize each network layer of EfficientSCI-S into 4-bit. Thus, we try to detect the source of this severe performance drop in the experiments. {\em First}, based on the 8-bit quantized baseline model which directly quantize each network layer into 8-bit, we set the bit-width of one module (the feature extraction module, ResDNet module or video reconstruction module) to 4-bit while maintaining that of the other two modules. As we can see in Tab.~\ref{Tab:feat}, there exists a 2.22 dB performance drop when we quantize the feature extraction module into 4-bit. However, the performance drop is relatively small with 0.52 dB and 0.49 dB for the 4-bit quantized ResDNet module and video reconstruction module. Therefore, we infer that the severe performance drop mainly comes from the low-bit quantized feature extraction module. {\em Stepping forward}, we visualize the output feature map of the feature extraction module to further verify our assumption. As shown in Fig.~\ref{fig:feat_map}, compared with full-precision model (Fig.~\ref{fig:feat_map}(a)), the output feature map of the feature extraction module in 4-bit quantized baseline model (Fig.~\ref{fig:feat_map}(b)) degrades severely. 
Therefore, we can conclude that the large performance drop of the low-bit quantized model mainly comes from the low-bit quantized feature extraction module.  

\subsection{\bf Proposed Quantization Framework}

\subsubsection{\bf Quantized Feature Extraction Module}
\label{sec:feat}

We have noticed from the empirical analysis (detailed in Sec.~\ref{sec:analysis}) that high-quality features play an important role in the performance improvement of the low-bit quantized model. Therefore, in this section, we present a simple and high-quality feature extraction module to improve the performance of the low-bit quantized  reconstruction methods. First, shortcut connections are widely used to propagate high-quality information. However, there exists a dimension mismatch problem during feature reshaping in the previous feature extraction module (shown in the red box of Fig.~\ref{fig:network}(b)). Thus, we propose to adopt several $1\times1\times1$ convolution layers as shortcut connections. Note that a pixel shuffle operation is performed before the last shortcut connected convolution layer to calibrate the spatial size mismatch. Then, we set the bit-width of the shortcut connected convolution layers to 8-bit to precisely propagate high-quality features. In this way, the proposed high-quality feature extraction module is capable of extracting and propagating high-quality features. Finally, when we compare Fig.~\ref{fig:feat_map}(b) and Fig.~\ref{fig:feat_map}(c), obvious quality improvement of the output feature map can be observed, which verifies the effectiveness of our proposed feature extraction module.

\begin{figure}[!t]
    \centering 
    \includegraphics[width=1.\linewidth]{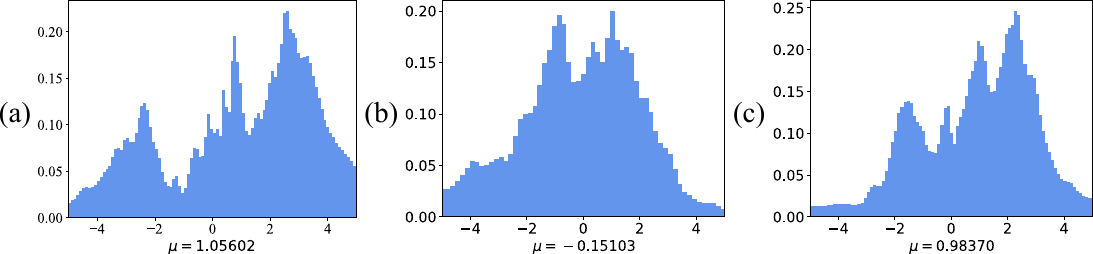}
    \caption{\small{The histogram of the query distribution in the first CFormer block of the first ResDNet block with different quantized models: (a) Full-precision model, (b) 8-bit quantized baseline model, and (c) Using the proposed shifted Transformer branch in 8-bit quantized baseline model.} 
    }
  \label{fig:q}
\end{figure}

\subsubsection{\bf Quantized Transformer Branch}

As shown in Q-ViT~\cite{li2022q}, there exists a distortion of the query and key distributions in the low-bit quantized vision Transformer branch, which brings performance drop. Therefore, we plot the distribution of the query activation in Fig.~\ref{fig:q} to better measure the distribution distortion of the low-bit quantized Transformer branch. We can see from Fig.~\ref{fig:q}(a) and Fig.~\ref{fig:q}(b) that the mean of the query distribution in the first CFormer block of the first ResDNet block between 8-bit quantized baseline model and its full-precision counterpart is 1.207 (1.056 $v.s.$ -0.151). Unfortunately, we cannot directly adapt Q-ViT into the reconstruction methods due to the absence of bell-shaped distributions as shown in Fig.~\ref{fig:q}(b). Therefore, we adapt a shift operation into the Transformer branch to rectify the query and key distributions. Specifically, given the query $\qv$ and key $\kv$, we  define a shift operation as $\widetilde{\qv}=\qv+\beta_{\qv}$ and $\widetilde{\kv}=\kv+\beta_{\kv}$, where $\beta_{\qv}$ and $\beta_{\kv}$ denote the learnable shift bias of the distributions. Finally, comparing Fig.~\ref{fig:q}(a) with Fig.~\ref{fig:q}(c), we observe that the mean of the query distribution in the first CFormer block of the first ResDNet block of 8-bit quantized baseline model equipped with our proposed shifted Transformer branch stays close to that of its full-precision counterpart (1.056 $v.s.$ 0.984), which verifies the effectiveness of the proposed approach. Due to the space limitation, please refer to the supplementary material for more distribution histograms of different CFormer blocks.   

\subsubsection{\bf Quantized Video Reconstruction Module}

In this section, to better propagate high-quality features through the network, we design a precise video reconstruction module to further bridge the performance gap with the following considerations: {\bf {i)}} As demonstrated in Sec.~\ref{sec:feat}, the proposed high-quality feature extraction module can extract and then propagate high-quality features to the ResDNet module. {\bf {ii)}} As illustrated in Fig.~\ref{fig:para_sota}(c), there are already some skip connections in the ResDNet module of EfficientSCI, which are able to propagate the high-quality features to the video reconstruction module. Therefore, we want to further propagate the high-quality features to the end of the network. As the video reconstruction module is also stacked with some convolution layers, we present a similar design with Sec.~\ref{sec:feat}. Specifically, as shown in Fig.~\ref{fig:network}(c), several $1\times1\times1$ shortcut connected convolution layers are added above the previous video reconstruction module (shown in the red box of Fig.~\ref{fig:network}(c)). Finally, we also set the bit-width of the shortcut connected convolution layers as 8-bit to ensure precise feature propagation.    

\subsection{\bf Architecture and Variants}

In this section, we choose EfficientSCI-S as a full-precision backbone model to customize quantized networks due to its superior efficiency performance. However, this does not mean that our proposed Q-SCI framework is equal to a low-bit quantized EfficientSCI-S, because our Q-SCI framework can also be applied on the other end-to-end video SCI reconstruction methods (See Tab.~\ref{Tab:stformer}). 

After that, we customize four quantized network variants including Q-SCI (8-bit), Q-SCI (4-bit), Q-SCI (3-bit), and Q-SCI (2-bit). The network quantization settings are as follows: {\bf {i)}} In the Q-SCI (8-bit) network, we directly set the bit-width of all the network layers to 8-bit and then adopt the proposed shifted Transformer branch. {\bf {ii)}} In the Q-SCI (4-bit), Q-SCI (3-bit), and Q-SCI (2-bit) network, we first set the bit-width of all the network layers to 4-bit, 3-bit, and 2-bit, respectively. Then, to ensure high-quality reconstruction results, we use our proposed high-quality feature extraction module, precise video reconstruction module and shifted Transformer branch. Note that previous work~\cite{esser2019learned} also sets a part of the network layers as 8-bit to ensure high performance. Additionally, since most layers of the low-bit quantized reconstruction networks are set to be (2, 3, or 4)-bit, it can still be considered as a (2, 3, or 4)-bit quantized model.

\subsection{\bf Loss Function}

Our proposed low-bit quantized network takes the measurement (\Ymat) and the corresponding masks ($\{\Mmat_t\}_{t=1}^{T}$) as inputs, and then generates the reconstructed video frames ($\{\hat{\Xmat}_t\}_{t=1}^{T} \in\mathbb{R}^{n_x\times{n_y}}$). To train the network, we choose the mean squared error (MSE) as our loss function,
\setlength\abovedisplayskip{3pt}
\setlength\belowdisplayskip{3pt}
\begin{equation}
\mathcal{L}_{MSE}=  \frac{1}{Tn_xn_y}\sum_{t = 1}^{T}\Vert \hat{\Xmat}_t - \Xmat_t \Vert_{2}^{2},
\end{equation}
where $\{\Xmat_t\}_{t=1}^{T}$ denotes the ground truth.

\section{Experimental Results}
\label{Sec:result}

\begin{figure}[!t]
    \centering 
    \includegraphics[width=1.\linewidth]{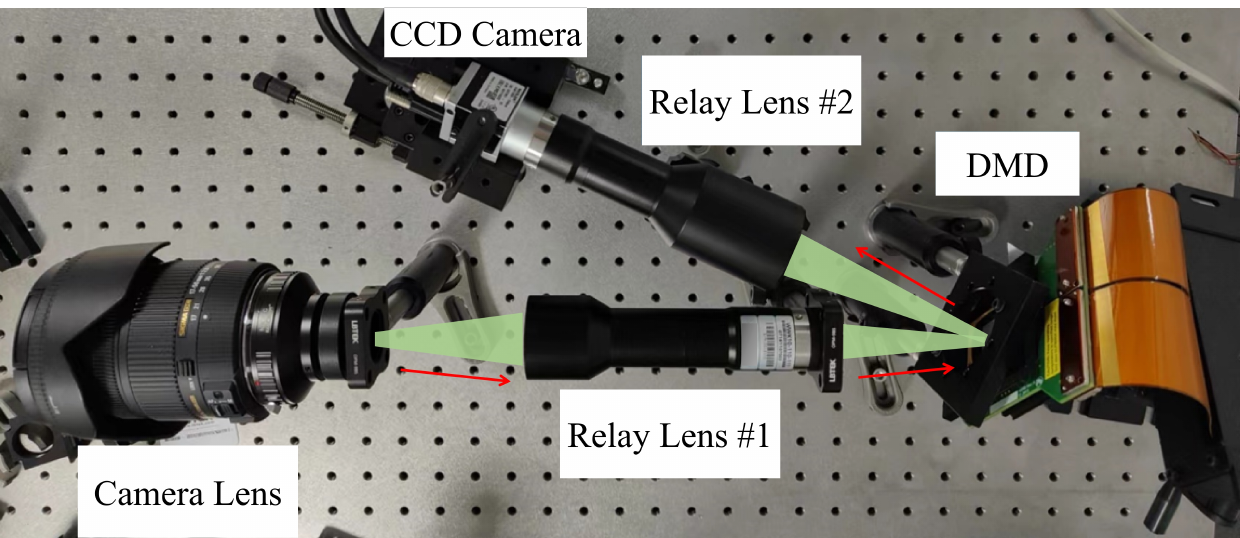}
        \caption{\small{Illustration of our real built video SCI system.
        } 
    }
  \label{fig:video-sm}
\end{figure}

\noindent{\bf Video SCI Hardware:} The optical setup of the real video SCI system is shown in Fig.~\ref{fig:video-sm}. The hardware encoding process of the video SCI system is as follows: First, the reflected light from the dynamic scene is imaged onto the surface of the digital micromirror device (DMD) via a camera lens (Sigma, 17-50/2.8, EX DC OS HSM) and the first relay lens (Coolens, WWK10-110-111). Following this, the projected high-speed scene on the DMD (TI, $2560 \times 1600$ pixels with 7.6 $\mu m$ pixel pitch) is modulated by the pre-stored random binary masks. Finally, the encoded compressed measurement is projected onto the surface of a low-cost CCD camera (Basler acA1920, $1920 \times 1200$ pixels with 4.8 $\mu m$ pitch pixel) with the second relay lens (Coolens, 
WWK066-110-110), which is captured in a single exposure time. Note that the CCD camera works at 50 FPS when capturing real data and the compression ratio of the video SCI system is Cr, then the equivalent sampling rate of our real video SCI system is 50$\times$Cr FPS.

\noindent{\bf Datasets:} Three types of dataset are used in this paper: training dataset, simulated testing dataset, and real testing dataset. {\bf {i)}} First, following BIRNAT~\cite{cheng2022recurrent}, we choose the \texttt{DAVIS2017} dataset~\cite{pont20172017} with spatial resolution $480\times{894}$ (480p) as the training dataset. {\bf {ii)}} We then test our Q-SCI framework on six simulated testing datasets, including \texttt{Kobe}, \texttt{Traffic}, \texttt{Runner}, \texttt{Drop}, \texttt{Crash}, and \texttt{Aerial} (with spatial resolution $256\times{256}$ and compression ratio is 8). For the simulated testing datasets, average Peak Signal-to-Noise-Ratio (PSNR) and average Structured Similarity Index Metrics (SSIM)~\cite{wang2004image} are used as the evaluation metrics of reconstruction quality. Besides, following previous works~\cite{cai2023binarized,qin2024quantsr,ma2023ompq}, model size (Params) and number of operations (OPs) are used as the evaluation metrics of algorithm efficiency. {\bf  {iii)}} Finally, we test the proposed Q-SCI framework on two real testing datasets, including \texttt{Domino} and \texttt{Water Balloon} (with spatial resolution $512\times{512}$ and compression ratio is 10) captured by our real built video SCI system described in the ``Video SCI Hardware'' section. Moreover, please refer to the supplementary material for more details about the real data capture process.

\noindent{\bf Training Process:} We build our models with PyTorch, and conduct training on 4 NVIDIA RTX 3090 GPUs. First, regular data augmentation operations such as random cropping, random flipping and random scaling are performed on the training dataset. Then, we initialize our Q-SCI with the full-precision EfficientSCI-S and adopt the Adam~\cite{kingma2015adam} optimizer to optimize the model with an initial learning rate of 0.0001. After iterating for 100 epochs on the training data with a 128$\times$128 spatial size and 20 epochs on the training data with a 256$\times$256 spatial size, we adjust the learning rate to 0.00001 and continue to iterate for 20 epochs on the training data with a 256$\times$256 spatial size. 

\begin{table*}[!t]
  \renewcommand{\arraystretch}{1.0}
  \caption{\small{The average PSNR in dB (left entry), SSIM (right entry), model size and computational cost of different reconstruction algorithms on six simulated testing datasets. 
  The best results are shown in bold and the second-best results are underlined.}}
  \centering
  \resizebox{\textwidth}{!}
  {
  \centering
  \begin{tabular}{l|cccccc|c|c|c}
  \toprule
  Method 
  & Kobe 
  & Traffic 
  & Runner 
  & Drop 
  & Crash 
  & Aerial 
  & Average 
  & Params (M) 
  & OPs (G) 
  \\
  \midrule
  MetaSCI~\cite{wang2021metasci} 
  & 30.12, 0.907
  & 26.95, 0.888
  & 37.02, 0.967
  & 40.61, 0.985
  & 27.33, 0.906
  & 28.31, 0.904
  & 31.72, 0.926
  & 2.07
  & 39.85
  \\
  BIRNAT~\cite{cheng2022recurrent} 
  & 32.71, 0.950
  & 29.33, 0.942 
  & 38.70, 0.976
  & 42.28, 0.992
  & 27.84, 0.927
  & 28.99, 0.917
  & 33.31, 0.951
  & 4.13
  & 390.56
  \\
  RevSCI~\cite{cheng2021memory}
  &33.72, 0.957
  &30.02, 0.949
  &39.40, 0.977
  & 42.93, 0.992
  & 28.12, 0.937
  & 29.35, 0.924
  & 33.92, 0.956
  & 5.66
  & 766.95
  \\
  STFormer-S~\cite{wang2022spatial}
  & 33.19, 0.955
  & 29.19, 0.941
  & 39.00, 0.979
  & 42.84, 0.992
  & 29.26, 0.950
  & 30.13, 0.934
  & 33.94, 0.958
  & 1.22
  & 193.47
  \\
  Dense3D-Unfolding~\cite{wu2021dense}
  & 35.00, 0.969
  & 31.76, 0.096
  & 40.03, 0.980
  & 44.96, 0.995
  & 29.33, 0.956
  & 30.46, 0.943
  & 35.26, 0.968
  & 61.91
  & 3975.83
  \\
  ELP-Unfolding~\cite{yang2022ensemble}
  & 34.41, 0.966
  & 31.58, 0.962
  & {41,16}, {0.986}
  & {44.99}, {0.995}
  & {29.65}, {0.959}
  & {30.68}, {0.944}
  & {35.41}, {0.969}
  & 565.73
  & 4634.94
  \\
  EfficientSCI-S~\cite{wang2023efficientsci}
  & 34.79, 0.968
  & 31.21, 0.961
  & {41.34}, {0.986}
  & 44.61, 0.994
  & {30.34}, {0.965}
  & {30.78}, {0.945}
  & \underline{35.51}, {\bf 0.970}
  & 3.78
  & 563.87
  \\
  \midrule
  Q-ViT (8-bit)~\cite{li2022q}
  & 34.60, 0.966
  & 30.65, 0.957
  & 40.94, 0.984
  & 43.91, 0.993
  & 30.14, 0.962
  & 30.75, 0.944
  & 35.17, 0.967
  & 0.95 
  & 141.04 
  \\
  Q-SCI (2-bit) (Ours)
  & 30.88, 0.920
  & 26.78, 0.899
  & 36.18, 0.957
  & 39.04, 0.975
  & 28.02, 0.917
  & 28.84, 0.901
  & 31.62, 0.928
  & {\bf 0.25} 
  & {\bf 19.85} 
  \\
  Q-SCI (3-bit) (Ours)
  & 32.93, 0.949
  & 29.08, 0.938
  & 38.62, 0.972
  & 41.70, 0.985
  & 29.25, 0.945
  & 30.10, 0.929
  & 33.62, 0.953
  & \underline{0.37} 
  & \underline{37.47} 
  \\
  Q-SCI (4-bit) (Ours)
  & 34.15, 0.962
  & 30.44, 0.954
  & 40.09, 0.980
  & 43.18, 0.990
  & 29.80, 0.956
  & 30.49, 0.938
  & 34.69, 0.963
  & 0.48 
  & 72.69 
  \\
  Q-SCI (8-bit) (Ours)
  & {34.95}, {0.968}
  & {31.24}, {0.961}
  & {41.60}, {0.985}
  & {44.27}, {0.993}
  & {30.34}, {0.963}
  & {31.03}, {0.945}
  & {\bf 35.57},  \underline{0.969}
  & 0.95 
  & 140.95 
  \\
  \bottomrule 
  \end{tabular}
  }
  \label{Tab:sim6}
\end{table*}

\begin{figure}[!ht]
    \centering 
        \includegraphics[width=.96\linewidth]{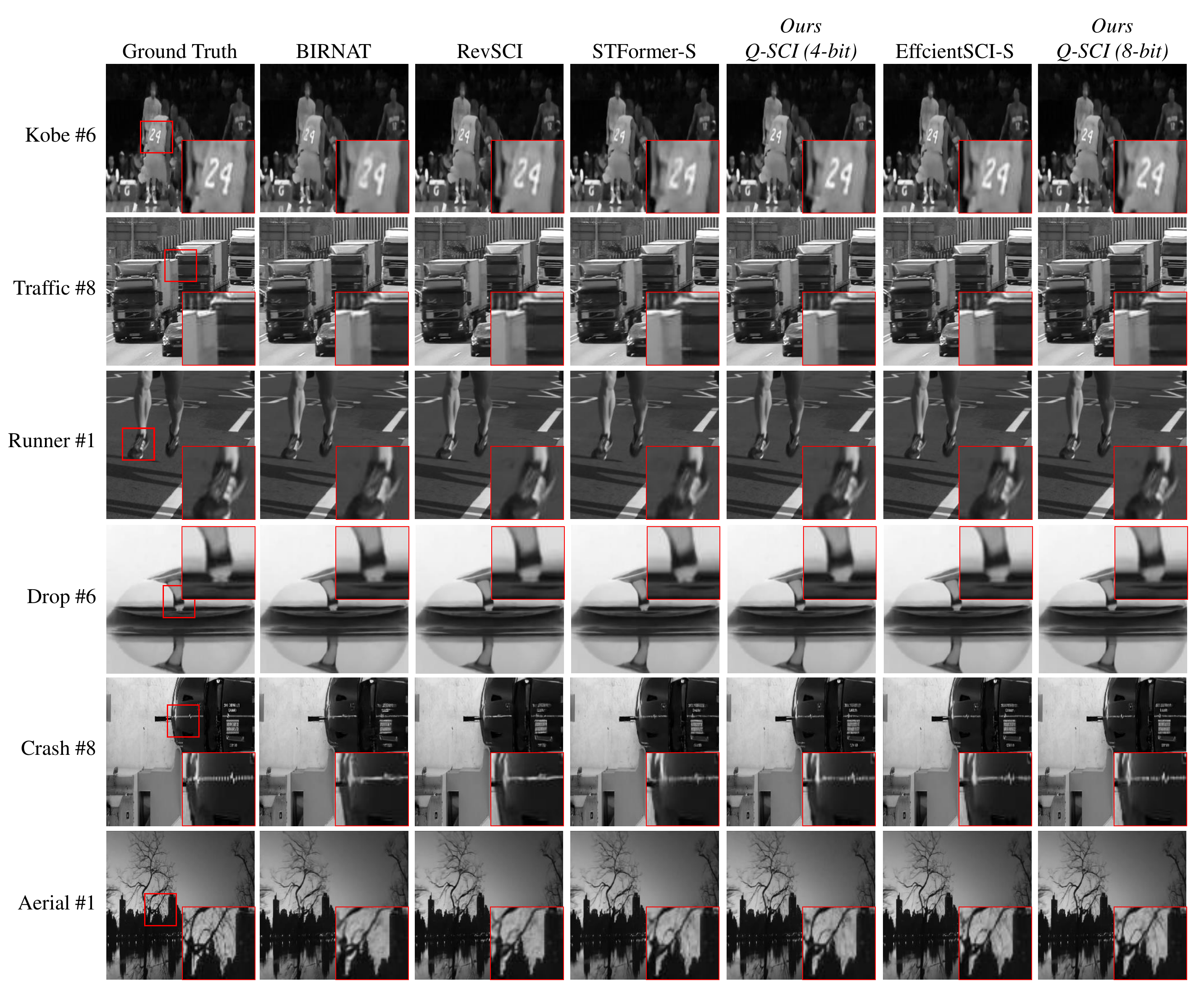}
    \caption{\small{Reconstructed video frames of the simulated testing datasets. For a better view, we zoom in on a local area as shown in the small red boxes of each ground truth image, and do not show the small red boxes again for simplicity. 
    }}
  \label{fig:sim}
\end{figure}

\subsection{Results on Simulated Testing Datasets}

In this section, we compare our Q-SCI networks with seven deep learning-based video SCI reconstruction methods (including MetaSCI, BIRNAT, RevSCI, Dense3D-Unfolding, ELP-Unfolding, STFormer-S, and EfficientSCI-S) on six simulated testing datasets. As shown in Tab.~\ref{Tab:sim6}, the proposed Q-SCI networks can achieve comparable reconstruction quality with {\em much fewer parameters and cheaper computational cost}. Specifically, {\bf {i)}} The proposed Q-SCI (2-bit) model achieves comparable reconstruction quality with MetaSCI, while the OPs is reduced by about 2$\times$. {\bf {ii)}} Our proposed Q-SCI (3-bit) model achieves 0.31 dB in PSNR higher than BIRNAT, and the OPs is reduced by about 10.4$\times$. {\bf  {iii)}} The proposed Q-SCI (4-bit) model achieves 1.38 dB and 0.77 dB in PSNR higher than BIRNAT and RevSCI, and the OPs is reduced by about 5.4$\times$ and 10.6$\times$. {\bf {iv)}} Our Q-SCI (8-bit) model can achieve 0.31 dB and 0.16 dB in PSNR higher than Dense3D-Unfolding and ELP-Unfolding, and the OPs is reduced by about 28.2$\times$ and 32.9$\times$. {\bf {v)}} Our proposed Q-SCI (8-bit) model achieves comparable performance with EfficientSCI-S, while the OPs is reduced by about 4$\times$. 

{For visualization purposes}, we also visualize some reconstructed video frames in Fig.~\ref{fig:sim}, where we can see from the zooming areas in each selected video frame that our proposed Q-SCI (4-bit) model can provide much clearer reconstructed images than BIRNAT, RevSCI, and STFormer-S. Further, the proposed Q-SCI (8-bit) model can provide comparable high-quality reconstructed images with EfficientSCI-S. Especially for the \texttt{Traffic}, \texttt{Crash}, and \texttt{Aerial} scenes, we can observe sharp edges and more details in the reconstructed frames of our proposed models, which verify the superior performance of the proposed Q-SCI quantization framework. 

Moreover, comparison with previous SOTA quantization method Q-ViT is also conducted.
Note that, we adapt both Q-ViT and Q-SCI into EfficientSCI-S for a fair comparison. The computational complexity and reconstruction quality are given in Tab.~\ref{Tab:sim6}. We can see from Tab.~\ref{Tab:sim6} that Q-SCI (8-bit) outperforms Q-ViT (8-bit) by about 0.4 dB with less computational cost. 

Finally, we adapt the proposed Q-SCI framework into previous SOTA end-to-end reconstruction method STFormer-S.
As shown in Tab.~\ref{Tab:stformer}, compared with 4-bit quantized baseline model, adapting our proposed high-quality feature extraction module will bring a 3.23 dB reconstruction quality improvement for STFormer-S. Then, further adapting the proposed precise video reconstruction module can lead to a 0.25 dB reconstruction quality improvement for STFormer-S. Therefore, the proposed Q-SCI framework can {\em well generalize} to other end-to-end video SCI reconstruction methods. 

\begin{figure}[!t]
    \centering 
    \includegraphics[width=.96\linewidth]{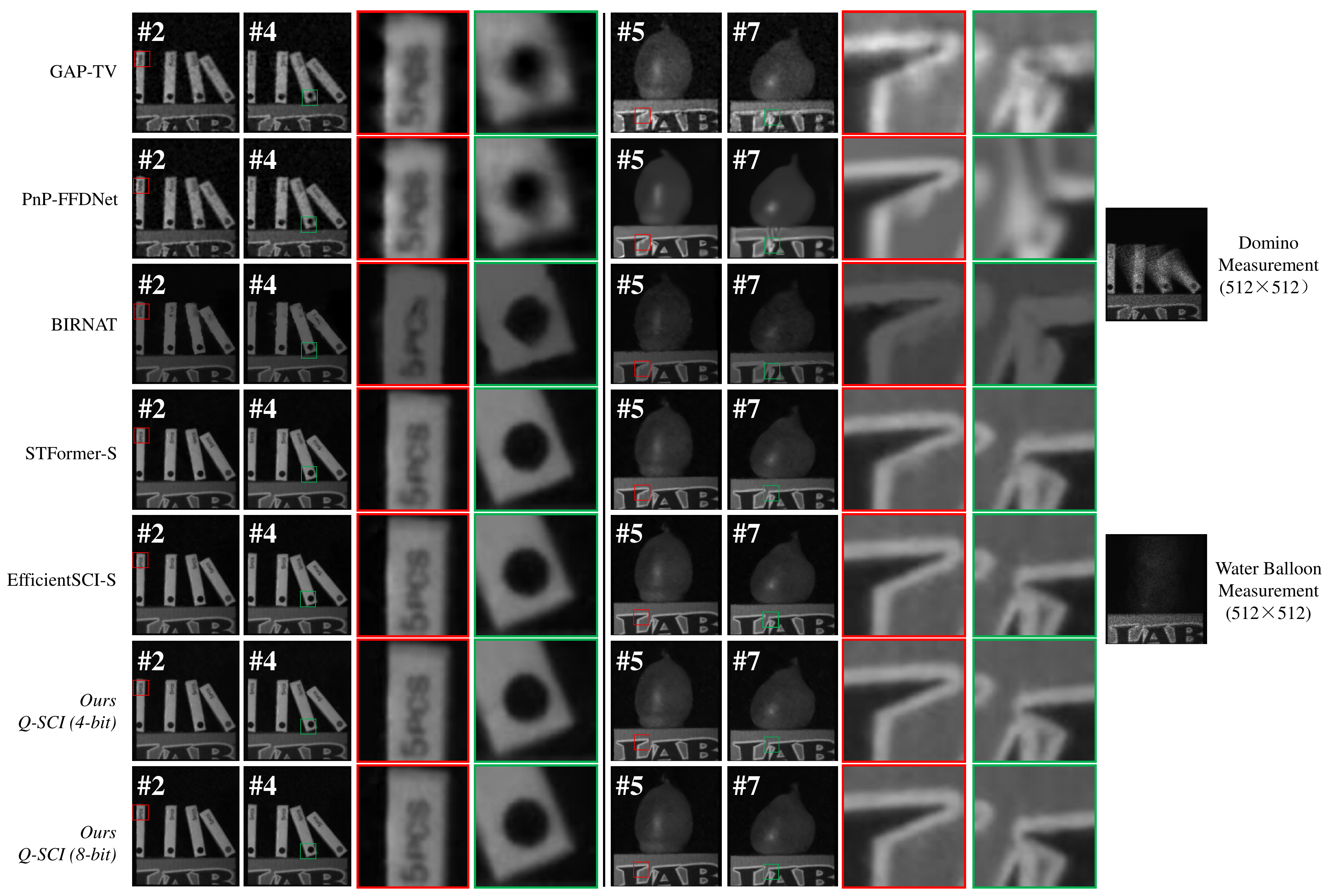}
    \caption{\small{Reconstructed video frames of the real data. 
    See \textcolor{blue}{Supplementary Movie 1} for the complete video. 
    }}
  \label{fig:real}
\end{figure}

\begin{table}[ht]
    \centering
    \begin{minipage}{0.49\linewidth}
        \centering
        \caption{\small{Verification on the {\em \bf generalization ability} of our proposed Q-SCI framework, where ``FEM'' and ``VRM'' stand for adapting the proposed high-quality feature extraction and precise video reconstruction modules into 4-bit baseline model of STFormer-S.}}
        \resizebox{.88\textwidth}{!}{
       \begin{tabular}{l|cc}
          \toprule
        Method & PSNR  & SSIM  
        \\
        \midrule
        4-bit Baseline & 30.03 & 0.903 \\
        4-bit Baseline+FEM & 33.26 & 0.949 \\
        4-bit Baseline+FEM+VRM & 33.51 & 0.952 \\
        \bottomrule
        \end{tabular}}
        \label{Tab:stformer}
    \end{minipage}\hfill
    \begin{minipage}{0.49\linewidth}
        \centering
        \caption{\small{Break-down ablation study towards higher performance, where ``RDM'', ``FEM'', and ``VRM'' represent using our shifted Transformer branch, high-quality feature extraction module, and precise video reconstruction module into 4-bit baseline model of EfficientSCI-S.}}
        \resizebox{1.\textwidth}{!}
        {
        \begin{tabular}{l|cc|cc}
          \toprule
          Method & PSNR &SSIM &Params(M) &OPs(G)
          \\
          \midrule
          4-bit Baseline  &31.40 &0.931  &0.473 &70.477
          \\
          +RDM &31.93 &0.929 &0.473 &70.477 
          \\
          +RDM+FEM &34.28 
          &0.959 
          &0.482 
          &71.857 
          \\
          +RDM+FEM+VRM &34.71  
          &0.963  
          &0.483
          &72.692
          \\
          \bottomrule
          \end{tabular}}
          \label{Tab:ablation}
    \end{minipage}
\end{table}

\subsection{\bf Results on Real Testing Datasets}

In this section, we test the proposed Q-SCI (8-bit) and Q-SCI (4-bit) models on the real testing datasets. Due to the uncertain noises of the real video SCI system, it is more difficult to reconstruct real measurements. As shown in Fig.~\ref{fig:real}, our proposed Q-SCI (4-bit) model can reconstruct clearer borders than previous real-valued models including GAP-TV, PnP-FFDNet, and BIRNAT. Moreover, we can clearly recognize the letters on the \texttt{Domino} on the reconstructed frames of the proposed Q-SCI (4-bit) and Q-SCI (8-bit). Finally, our proposed Q-SCI (8-bit) model can present sharper boarders and easier-recognized letters on the \texttt{Domino} when compared with the real-valued STFormer-S and EfficientSCI-S.  

\subsection{Ablation Study}

Here, we conduct a break-down ablation towards higher performance. We first build a baseline method by directly quantizing each layer of EfficientSCI-S into 4-bit. As shown in Tab.~\ref{Tab:ablation}, the baseline model yields 31.40 dB in PSNR and 0.931 in SSIM. When we adopt the shifted Transformer branch, the model achieves a 0.53 dB improvement (which is a moderately large improvement in the video SCI reconstruction task) with neglect computational burden. After that, we apply the high-quality feature extraction module and the precise video reconstruction module successively. The performance improvement are 2.35 dB and 0.43 dB, respectively. Finally, we obtain the Q-SCI (4-bit) model with 34.71 dB which is a totally 3.31 dB improvement against the 4-bit quantized baseline model, while the computational cost only grows by 3.14$\%$. 
These experimental results verify the effectiveness of the proposed module designs of our Q-SCI framework.

\section{\bf Conclusion and Future Work}
\label{Sec:conclude}

In this paper, we propose the first low-bit quantization framework for the end-to-end video SCI reconstruction methods, which is {\em simple and effective}.
Specifically, we design a high-quality feature extraction module and a precise video reconstruction module to extract and propagate high-quality features in the low-bit quantized model. Additionally, we introduce a shift operation on the query and key distributions of the low-bit quantized Transformer branch to further bridge the performance gap. Finally, we verify that our proposed low-bit quantization framework can generalize well to different end-to-end reconstruction methods (including EfficientSCI and STFormer). 

While our proposed Q-SCI can largely reduce computational cost with small performance drop, there is still room for further improvement, which we discuss as follows. First of all, deploying Q-SCI on the resource-limited devices (\ie, smart phone and autonomous vehicle)  requires extensive engineering and hardware support, making it a challenging task. Secondly, although Q-SCI is a well-designed network quantization framework for video SCI, a novel mobile-friendly reconstruction network will be more suitable than previous reconstruction methods for the quantized video SCI reconstruction task. Finally, algorithm-hardware codesign should be considered to better fit the resource-limited devices and the video SCI hardware encoding system.  

Regarding the future work, we can further explore other model compression methods including network pruning~\cite{wang2022trainability,li2022revisiting} and knowledge distillation~\cite{kordopatis2022dns,wang2022makes} to further optimize the efficiency performance. Moreover, we expect to extend the proposed Q-SCI framework to various imaging systems such as high-speed imaging~\cite{luo2024snapshot,dou2023coded,zheng2023capture}, hyperspectral imaging~\cite{fang2024wide,lin2023metasurface,meng2020end}, depth of field imaging~\cite{chen2023meta,yuan2014low,shen2023monocular}, and single-pixel imaging~\cite{wang2023mid,qu2024dual,xu2024compressive}.

\section*{Acknowledgements}
This work was supported by the National Natural Science Foundation of China (grant number 62271414), Zhejiang Provincial Distinguished Young Scientist Foundation (grant number LR23F010001), Zhejiang ``Pioneer” and ``Leading Goose” R\&D Program  (grant number 2024SDXHDX0006, 2024C03182), the Key
Project of Westlake Institute for Optoelectronics (grant number 2023GD007), and
the 2023 International Sci-tech Cooperation Projects under the purview of the ``Innovation Yongjiang 2035” Key R\&D Program  (grant number 2024Z126). 

%
%
\bibliographystyle{splncs04}
\bibliography{main}
\end{document}